\documentclass[10pt,twocolumn,letterpaper]{article}

\usepackage[pagenumbers]{cvpr} % To force page numbers, e.g. for an arXiv version
\usepackage{float}

\usepackage[dvipsnames]{xcolor}

\definecolor{cvprblue}{rgb}{0.21,0.49,0.74}
\usepackage[pagebackref,breaklinks,colorlinks,citecolor=cvprblue]{hyperref}

\title{MonoGaussianAvatar: Monocular Gaussian Point-based Head Avatar}

\author{Yufan Chen$^{1\dag}$, Lizhen Wang$^2$, Qijing Li$^2$, Hongjiang Xiao$^3$, Shengping Zhang$^1$,\\ Hongxun Yao$^1$, Yebin Liu$^2$\\
$^1$ Harbin Institute of Technology $^2$ Tsinghua University $^3$ Communication University of China\\
{\small\url{yufan1012.github.io/MonoGaussianAvatar}}
}

\begin{document}
\twocolumn[{%
\maketitle
\begin{figure}[H]
\hsize=\textwidth %
\centering
\includegraphics[width=\textwidth]{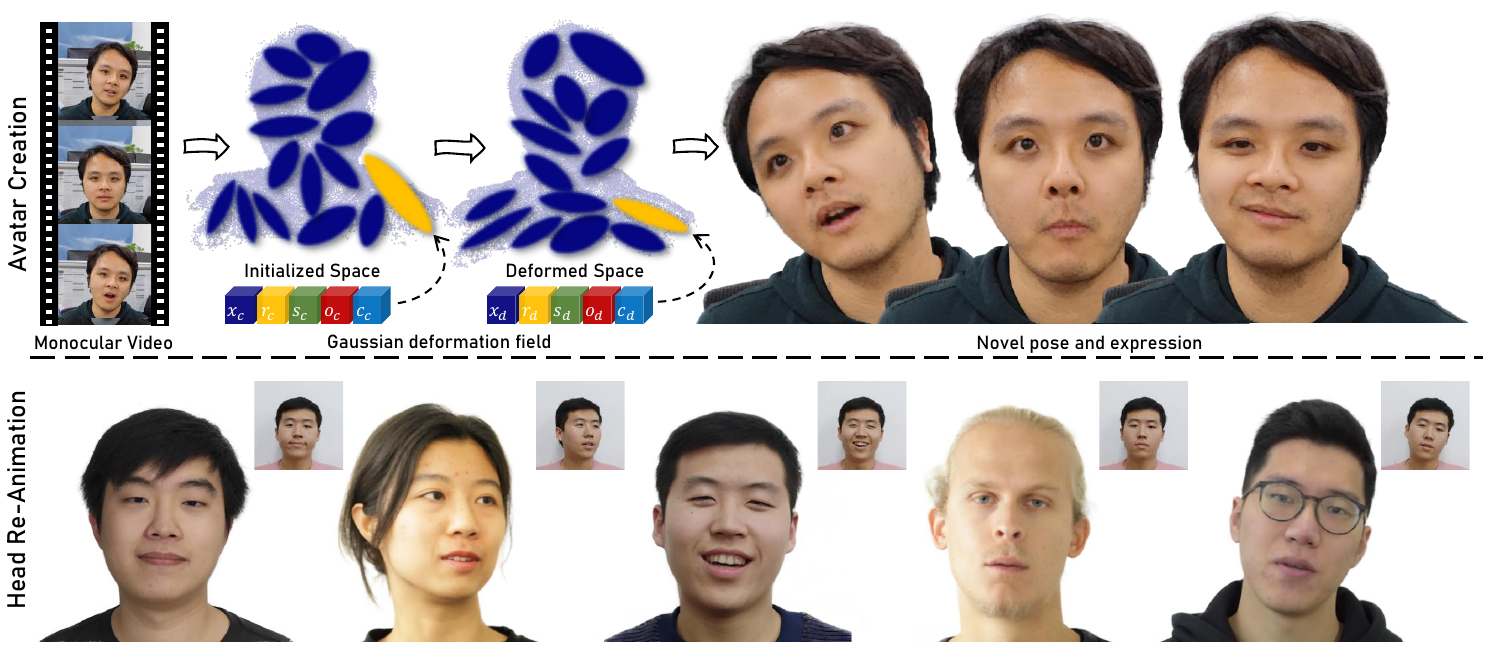}
\vspace{-0.6cm}
\caption{\textbf{MonoGaussianAvatr}
    reconstructs a dynamic facial avatar by applying a Gaussian deformation field from a monocular portrait video of a person, which allows us to synthesize novel poses and expressions.
    }
\label{fig: teaser}
\end{figure}
}]

\maketitle
\let\thefootnote\relax\footnotetext{$^\dag$ Work done during an internship at Tsinghua University.}
%%%%%%%%% ABSTRACT
\begin{abstract}

\end{abstract}
\vspace{-0.7cm}
The ability to animate photo-realistic head avatars reconstructed from monocular portrait video sequences represents a crucial step in bridging the gap between the virtual and real worlds.
Recent advancements in head avatar techniques, including explicit 3D morphable meshes (3DMM), point clouds, and neural implicit representation have been exploited for this ongoing research.
However, 3DMM-based methods are constrained by their fixed topologies, point-based approaches suffer from a heavy training burden due to the extensive quantity of points involved, and the last ones suffer from limitations in deformation flexibility and rendering efficiency.
In response to these challenges, we propose MonoGaussianAvatar (Monocular Gaussian Point-based Head Avatar), a novel approach that harnesses 3D Gaussian point representation coupled with a Gaussian deformation field to learn explicit head avatars from monocular portrait videos.
%
% We define our Gaussian points with Gaussian parameters to have flexible shapes, enabling flexible topology.
We define our head avatars with Gaussian points characterized by adaptable shapes, enabling flexible topology.
These points exhibit movement with a Gaussian deformation field in alignment with the target pose and expression of a person, facilitating efficient deformation. 
Additionally, the Gaussian points have controllable shape, size, color, and opacity combined with Gaussian splatting, allowing for efficient training and rendering.
Experiments demonstrate the superior performance of our method, which achieves state-of-the-art results among previous methods.
% MonoGaussianAvatar offers a unique blend of advantages, combining high-quality geometry, flexible topology, and efficient deformation and rendering capabilities through 3D Gaussian point representation.
% %
% Furthermore, it benefits from high-quality rendering results enabled by Gaussian splitting.
% %
% Experiments demonstrate the superior performance of our method, which achieves state-of-the-art results among previous methods.

%%%%%%%%% BODY TEXT
\section{Introduction}
\label{sec:introduction}
The urgent demand for automated construction of personalized 3D head avatars from monocular portraits is stressed by the evolving landscape of AR and VR wearables, such as Apple Vision Pro, Meta Quest, etc.
These avatars must meet the demanding criteria of photorealistic facial geometry, intricate facial appearance, and precise representation of dynamic expressions to fulfill the expectations of users.
Recent efforts have pursued three primary presentations for the creation of 3D head avatars from portrait videos: the 3D Morphable Model (3DMM), Neural implicit representation, and other explicit representations, such as PointAvatar.
3DMMs~\cite{blanz2023morphable, li2017learning, paysan20093d} excel in efficiently representing rough facial geometry and have the capacity to generalize to novel deformations. However, they~\cite{grassal2022neural, Khakhulin2022ROME} are limited by their predefined fixed topologies and are constrained to surface-like geometries, making them less suitable for modeling features such as eyeglasses, hair, and other accessories.
%这段需要修改 citation
Neural implicit representations~\cite{mescheder2019occupancy, park2019deepsdf, mildenhall2021nerf} can be categorized into NeRF-based~\cite{gafni2021dynamic, guo2021ad, liu2022semantic, gafni2021dynamic, athar2021flame, athar2022rignerf, gao2022reconstructing, bai2023learning, zielonka2023instant, zhao2023havatar, Gao2022nerfblendshape, xu2023avatarmav, xu2023latentavatar, duan2023bakedavatar} and SDF-based methods~\cite{zheng2022avatar}. The former strengthens in capturing intricate details such as hair strands and eyeglasses, while the latter demonstrates strength in reconstructing geometry. 
Nonetheless, their efficiency in training and rendering is compromised, as generating a single pixel necessitates querying numerous points along the camera ray.
For other explicit representations~\cite{bharadwaj2023flare}, PointAvatar~\cite{zheng2023pointavatar} stands out as an exemplary model showcasing superior performance compared to other methods.
Compared to implicit representations, points can be rendered efficiently with a standard differentiable rasterizer and deformed efficiently through Linear Blend skinning (LBS).
Compared to meshes, they are considerably more flexible and versatile, which confirms the topology of accessories and complex volumetric structures.
However, PointAvatar fixes its rendering radius uniformly which increases the number of required training points, imposing a heavier training burden.
%voids
It also results in a number of holes when the face undergoes substantial movement.
Moreover, certain rigid parts such as teeth may undergo similar deformation as the lips, leading to inaccuracies.
%加一句话

%
In this paper, to address the challenges associated with PointAvatar, we introduce Monocular Gaussian Point-based Head Avatar (MonoGaussianAvatar), as depicted in Fig.~\ref{fig: teaser}. We utilize 3D Gaussian points to represent the facial appearance and geometry while integrating a continuous Gaussian deformation field for animation.
%改一下
In contrast to the potential holes and inaccuracies introduced in rigid regions, such as teeth, Gaussian points demonstrate enhanced flexibility in scale and rotation. 
% This flexibility effectively compensates for the limitations inherent in traditional point rendering techniques.
%和上句话并在一块
Simultaneously,  Gaussian points exhibit superior rendering quality compared to PointAvatar when employing an equivalent number of points. This efficiency facilitates the allocation of additional Gaussian points, thereby further enhancing rendering details.
% 
% or rendering an image with higher resolution.
%删去， 找一句话，说下更高分辨率
% In contrast to point rendering in PointAvatar, where the rendering quality is directly influenced by the geometry, Gaussian splatting can leverage its anisotropy to compensate for limitations in geometry during rendering. 
%保留牙齿这句hole的优势，利用更少的点去渲染更高分辨率的图像
% Furthermore, the Gaussian representation enables the restoration of holes and the rigidity of structures, such as teeth.
%

%但是现有的高斯要么是静态，要是dynamic（heavy）的方法，目前还没有针对人脸单视角的高斯方法，同时直接把高斯直接应用在人脸任务上，也不是简单
% Existing Gaussian splatting methods are generally categorized as either static or dynamic.
% %
% Dynamic methods often involve frame-to-frame supervised tracking of points, which can be computationally intensive.
% %
% As of now, there is a noticeable absence of Gaussian-based methods utilizing monocular video for avatar creation.
% %
% Simultaneously, the direct application of Gaussian splatting to the task of monocular avatar creation presents inherent challenges and is not straightforward.
% %
% % One strength of our approach lies in introducing 3D Gaussian representation, enabling the simultaneous constraint of both the geometry and appearance of avatars.
% %动态人脸任务，高斯需要很好的初始化才能做好动态同拓扑，同时高斯额外的参数也需要在初始化阶段学习出来
% In dynamic facial tasks, Gaussian points require a solid initialization to consistently produce a dynamic topology. 

%
Although Gaussian representation can address the aforementioned challenges, existing Gaussian splatting methods face difficulties when applied to head avatar work.
To initialize and update the Gaussian parameters during training convergence, we employ a two-stage initialization strategy for Gaussians.
% In contrast to 3D-GS, which relies on iteration optimization for parameter updates, we propose a Gaussian prediction network to predict Gaussian parameters in canonical space.
%同时高斯原有的增删点策略无法适用于人脸avatar任务，以及高斯具有额外的参数需要跟随人脸动态表情变化
% Moreover, original 3D-GS is limited to reconstructing static scenes. When applied to dynamic scenes, it encounters issues of geometry inconsistency and appearance blur. Therefore, we introduce a Gaussian deformation field to facilitate a seamless transition from static to dynamic representations.
Moreover, the strategy of point insertion and deletion in 3D-GS fails to meet the requirements of head avatar tasks.
%于是，我们受pointavatar的启发，我们才用了一个两阶段的初始化策略，并且提出了一种更适用于人脸任务的增删点策略，针对于高斯的额外的参数我们提出了一个deformation field（变化的是offset）使这些参数能够跟随表情变化，由于高斯整体的训练流程需要点和图像对的比较齐，防止多帧的信息在同一帧被平滑，补充说明这个deformation是必要的
% Finally, the point insertion and deletion strategy in 3D-GS is based on adaptive density which is contingent on the backward gradient of iteration optimization and is not aligned with the process of creating an avatar.
Therefore, drawing inspiration from PointAvatar, we introduce a more fitting and innovative approach to point insertion and deletion for our method.
In addressing the adjustment of the Gaussian parameters with deformed expressions, we propose a Gaussian deformation field. 
Besides, given that the entire training process requires optimal alignment between points and images, the introduction of this Gaussian deformation field becomes imperative to prevent the smoothing of information from multiple frames into a single frame.

\medskip
\noindent
 
The contributions of this work are as follows:

\begin{itemize}
    %提一下快速渲染高清图像
    \item We first present a Gaussian point-based explicit head avatar model which is combined with facial expressions and poses that can reconstruct detailed geometry and appearance from monocular videos. 
    %
    % Besides, our model achieves a high-resolution rendering.
    %
    \item We introduce a Gaussian points insertion and deletion strategy,
    % which significantly enhances training stability and improves convergence efficiency.
    %
    Moreover, we employ the Gaussian deformation field to preserve the structure of accessories, such as glasses, in a novel pose.
    \item Our method demonstrates superior results compared with the SOTA methods in the metrics of structure similarity, image similarity, and Peak Signal-to-Noise Ratio.
    
    % \item We present an improvement of the Gaussian points setting, the optimization of Gaussian parameters, and the Gaussian deformation field regarding 3D Gaussian splitting, which shows state-of-the-art photo realism while achieving real-time rendering.

    %\item Our method recovers detailed wrinkles of head avatars just on monocular videos, which also achieves real-time rendering.
    % \item The framework achieves real-time rendering on human avatars, up to 30 FPS at a resolution of $1024 \times 1024$, while containing comparable performance with state-of-the-art methods.
\end{itemize}
% efficient 4D Gaussian representation by modeling the Gaussian parameters prediction field and Gaussian deformation field, which can be combined with facial expressions and poses to reconstruct 3D head avatars from monocular videos.

 % We introduce a set of point insertion and deletion strategies and, with the assistance of Gaussian rendering, enable the reconstruction of a dense point cloud avatar of a human head from a randomly initialized sparse point cloud.
\section{Related Work}
\begin{figure*}
    \centering
    \includegraphics[width=\textwidth]{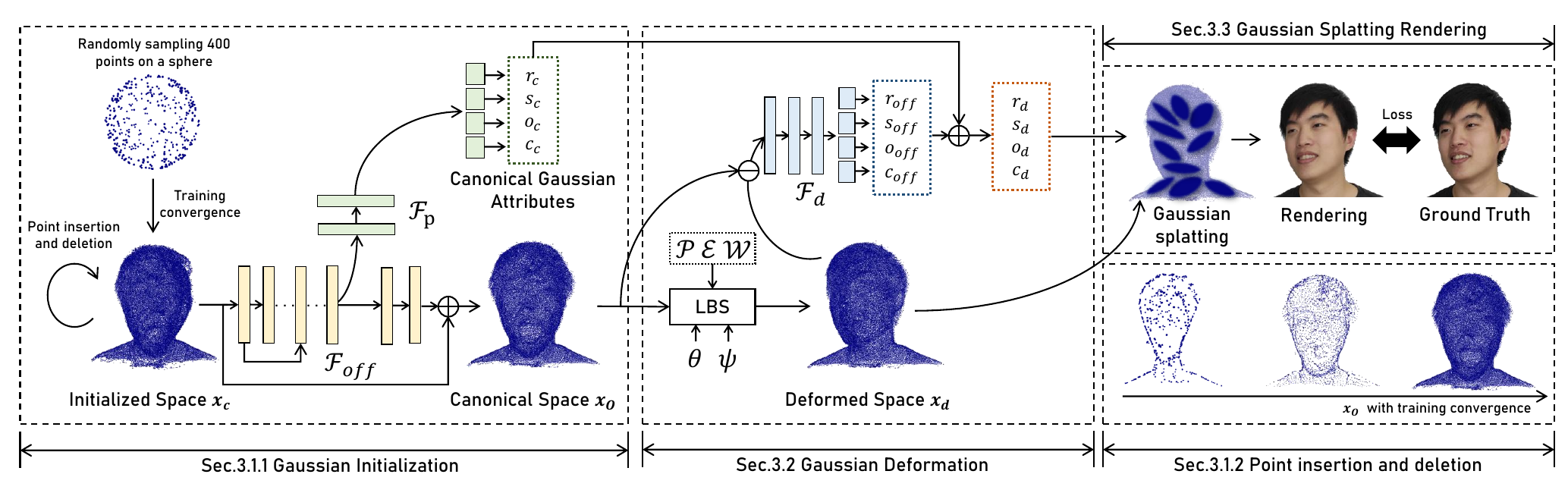}
    \caption{\textbf{Method Pipeline}.
    We model the human head as a learned parametric deformed 3D Gaussian points, comprising the mean position $x_d$, the color ${c}_{d}$, the opacity ${o}_{d}$, the rotation ${r}_{d}$, and the scale ${s}_{d}$. These parameters collectively characterize the subject's geometric and intrinsic appearance in the deformed space.
    The left module (Sec.~\ref{sec: Gaussian Initialization}) details the initialization process of our Gaussian representation, explaining how we obtain the initial mean position ${x}_{c}$, derive the other Gaussian parameters, and transform the mean position ${x}_{c}$ from the initialized space to the canonical space.
    %
    % Given a target expression and pose, we warp $x_c$ with a learned offset $\mathcal{O}$ to $x_o$ in FLAME canonical space. Subsequently, we deform them to $x_d$ in the target deformed space by applying blendshape and skinning, utilizing learned personalized blendshapes bases ($\mathcal{E}$, $\mathcal{P}$) and skinning weights ($\mathcal{W}$).
    % %
    The middle module (Sec.~\ref{sec: Gaussian deformation}) introduces the transformation process of the mean position ${x}_{o}$ from the canonical space to the deformed space using LBS. It also describes how the other Gaussian parameters in the deformed space are adjusted to fit the transformation of the mean position through the Gaussian parameter deformation field.
    %
    % to establish the relationship between $x_c$ and the other four parameters. Additionally, a Gaussian deformation field warps $\mathbf{\alpha_c}$, $\mathbf{r_c}$, and $\mathbf{s_c}$ to deformed space with their learned corresponding offset $\mathcal{O}_{\mathbf{\alpha}}, \mathcal{O}_{\mathbf{r}}, \mathcal{O}_{\mathbf{s}}$.
    %
    The right-top module (Sec.~\ref{sec: Gaussian Splatting Rendering}) presents the rendering process with 3D Gaussian parameters in the deformed space.
    The right-bottom (Sec.~\ref{sec: Point Insertion and Deletion}) demonstrates the strategy of point insertion and deletion.
    }
    \label{fig: pipeline}
\vspace{-0.5cm}
\end{figure*}
\paragraph{3D Head Avatar Reconstruction.} 
The reconstruction of 3D head avatars has been a challenging and widely studied task in the field of computer vision. Some works utilize the 3D morphable model (3DMM)~\cite{blanz2023morphable, li2017learning, paysan20093d} to represent the basic facial geometry and expressions. 
Neural Head Avatar~\cite{grassal2022neural} employs two feed-forward networks to predict vertex offsets and textures, enabling the extrapolation to unseen facial expressions. 
ROME~\cite{Khakhulin2022ROME} uses a neural texture map to improve the quality of rendering images. However, these methods still fail to represent specific components like hair because of the limitation of the fixed topologies. 

The success of neural implicit representation~\cite{mescheder2019occupancy, park2019deepsdf, chen2019learning, yariv2020multiview} has opened up new possibilities for creating 3D head avatars. 
IMAvatar~\cite{zheng2022avatar} proposes the utilization of signed distance functions (SDFs) to depict an implicit head model, which enables accurate geometry.
Nerface~\cite{gafni2021dynamic} first introduces a 4D facial avatar model based on NeRF~\cite{mildenhall2021nerf}.
A variety of methods utilize powerful techniques to model a NeRF-based avatar~\cite{guo2021ad, liu2022semantic, gafni2021dynamic, athar2021flame, athar2022rignerf, gao2022reconstructing}, such as combining UV maps and local NeRF~\cite{bai2023learning}, Instant-NGP~\cite{zielonka2023instant}, triplane~\cite{zhao2023havatar}, semantic blendshape coefficients~\cite{Gao2022nerfblendshape}, prior information of the face~\cite{xu2023avatarmav}, and Latent Expression Code~\cite{xu2023latentavatar}, deformable layered meshes~\cite{duan2023bakedavatar}, enhancing the training speed or improve the quality of rendered images.
Nonetheless, these methods still face challenges in terms of efficiency during both training and rendering processes.

Therefore, some works utilize explicit representations to improve efficiency. FLARE~\cite{bharadwaj2023flare} employs a face avatar model based on mesh to facilitate efficient rasterization. However, mesh representations are limited in their ability to capture fine structures such as hair. 
Point clouds have been considered as an alternative with a more flexible topology and easier deformability. PointAvatar~\cite{zheng2023pointavatar} generates a point-based face model by optimizing the deformation of point clouds. This approach allows for the rendering of high-quality images and provides accurate representations of challenging components of the head, such as eyeglasses and hair. By optimizing the deformation of the point clouds, PointAvatar achieves controllable animations, resulting in realistic and expressive facial animations. However, the fixed rendering radius of normal point clouds presents a limitation as it needs to strike a balance between rendering quality and the number of points. When the face undergoes substantial movement, PointAvatar encounters several rendering issues, leading to the presence of holes in the output.

Traditional approaches~\cite{levoy2000the, beeler2010high, ghosh2011multiview, bradley2010high, lombardi2018deep, lombardi2021mixture, ma2021pixel, wang2023neural} are capable of modeling full-head avatars from dense multi-view videos, allowing for the reconstruction of accurate geometry and the achievement of high-fidelity animation. However, these methods are often limited by their heavy computation.

\paragraph{3D Gaussian Splatting.}
A new approach called 3D-GS~\cite{kerbl20233d} has emerged in point cloud rendering. It has demonstrated success in new view synthesis, rapid training speed, and real-time rendering. In comparison to the traditional point rendering approach, Gaussian points provide several advantages. They not only maintain the explicit representation of traditional points but also offer improved flexibility in terms of scale and rotation. This enhanced flexibility makes Gaussian points particularly suitable for modeling rigid regions in the face, such as teeth. By leveraging Gaussian points, it is possible to compensate for the presence of gaps or holes that may occur when using traditional point rendering methods. Recently, the incorporation of 3D-GS in tracking~\cite{luiten2023dynamic}, 3D content generation~\cite{tang2023dreamgaussian} and dynamic reconstruction~\cite{xu20234k4d, wu20234dgaussians, yang2023deformable3dgs} has demonstrated significant potential and capabilities. Our work draws inspiration from these developments and implements the use of 3D Gaussian points in head avatars. This integration has led to high-resolution rendering and has demonstrated excellent performance in handling novel poses and structures, including glasses.

\section{Method}
\label{sec:method}
We propose MonoGaussianAvatar, a Gaussian point-based morphable head avatar enabling 4D reconstruction using a monocular portrait video of a subject performing diverse expressions and poses.
We start by defining Gaussian points and incorporating a point insertion and deletion strategy. This strategy dynamically updates the mean position of Gaussian points with training convergence in the initialized space.
Subsequently, a Gaussian parameter prediction network is employed to generate the Gaussian parameters (opacity, rotation, scale, and color) based on the mean position of each Gaussian point in the initialized space.
The mean positions of the Gaussian are then deformed from the initialized space to the canonical space utilizing a learnable offset. The FLAME expression and pose parameters determine the mean position of the Gaussian in the deformed space.
Finally, rendering is achieved through 3D Gaussian splitting, involving both the Gaussian parameters obtained from the Gaussian parameter prediction network in the initialized space and the Gaussian deformation field in the deformed space.
These components collaboratively optimize the rendering of 3D Gaussian points in deformed space, aligning with the ground truth images and ensuring a faithful representation of the subject's expressions and poses.
The schematic overview of our method is depicted in Fig.~\ref{fig: pipeline}.
\subsection{Gaussian Setting}
\label{sec: Gaussian Setting}
We introduce a two-stage initialization strategy and a more fitting and innovative approach to point insertion and deletion for our avatar work.
\subsubsection{Gaussian Initialization}
\label{sec: Gaussian Initialization}
%参数预测网络还没讲
In this section, we introduce the attributes of our 3D Gaussian representation in initialized space, including learnable mean position $x_c\in{\mathbb{R}^3}$, rotation $r_c{\in}{\mathbb{R}^4}$, scale $s_c{\in}{\mathbb{R}^3}$, opacity $o_c{\in}{\mathbb{R}}$ and color $c_c{\in}{\mathbb{R}^3}$.
Therefore, our Initialized Gaussian points can be defined as follows
\begin{equation}
\label{eq:canonical_gaussian_points}
    \mathcal{{G}}_{c} =\{{x_c}^{i}, {r_c}^{i}, {{s_c}^i}, {o_c}^{i}, {{c_c}^i}\}_{i=1:N} 
\end{equation}
In the first stage, we initialize the mean position ${x}_{c}$ by randomly sampling 400 points on a sphere, As the training progresses, we employ the strategy of point insertion and deletion to iteratively update its value, as detailed in Sec.~\ref{sec: Point Insertion and Deletion}.
Sequentially, the mean position is utilized to calculate the opacity, rotation, and scale of the Gaussian point, along with the color.
To facilitate this process, we introduce a Gaussian parameter prediction network, denoted as ${\mathcal{F}}_{\rm p}(\cdot)$ employing an MLP to map the mean position $x_c$ to the rotation $r_c$, scale $s_c$, opacity $o_c$ and color $c_c$ in the initialized space
\begin{equation}
\label{eq:initialized_gaussian_parameters}
    (r_c, s_c, o_c, c_c)  = {\mathcal{F}}_{\rm p}(x_c)
\end{equation}
Then, we warp the Gaussian point from the initialized space to canonical space, which corresponds to the FLAME~\cite{li2017learning} temple with a predefined mouth-opened pose, in preparation for deformation.
The Gaussian points in canonical space have their positions updated with a learnable function $\mathcal{F}_{off}(\cdot)$ mapping the offset from initialized space to canonical space
\begin{equation}
\label{eq:initialized_canonical__gaussian_parameters}
    x_o=x_c + \mathcal{F}_{off}(x_c)
\end{equation}
\subsubsection{Point Insertion and Deletion}
\label{sec: Point Insertion and Deletion}
\begin{figure}
    \centering
    \includegraphics[width=0.47\textwidth]{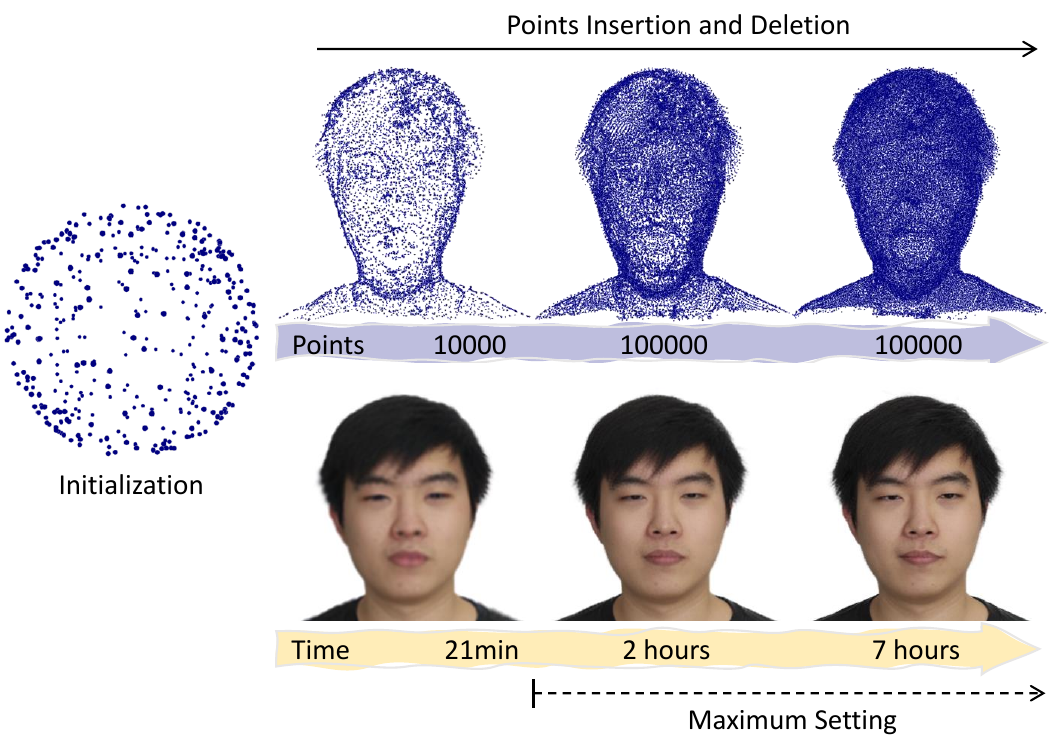}
    \vspace{-0.1cm}
    \caption{The strategy of point insertion and deletion intricately outlines the final reconstruction process, encompassing both appearance and geometry. It facilitates rapid convergence and has the capacity for refinement when the number of points reaches the maximum setting.
    }
    \label{fig: point_insertion}
\vspace{-0.5cm}
\end{figure}
In Sec.~\ref{sec: Gaussian Initialization}, we initialize the mean position ${x}_{c}$ of the Gaussian point in initialized space to calculate the other Gaussian parameters. However, the initialized Gaussian point can only aid in reconstructing a coarse geometry with $\mathcal{F}_{off}$. Therefore, it is essential to design a strategy for point insertion and deletion to achieve a coarse-to-fine process.
Our strategy of point insertion and deletion is similar to that of PointAvatar which prunes away invisible points that are not the first points splatting on each pixel when rendering with an epoch-decay uniform radius every epoch and doubles the number of points with the same sampling radius every 5 epochs.
However, 3D Gaussian splitting (see Sec.~\ref{sec: Gaussian Splatting Rendering}) has its splatting scale various and also rotation and opacity controlling shape and transmittance, which makes rendering more flexible and efficient as well as more challenging to maintain the stability of the training process compared to that of PointAvatat.
We attempted to replicate the PointAvatar procedure, but the rendering result turned out to be coarse, and the geometry appeared rough (see the experiment in Sec.~\ref{sec: Ablation Point Insertion and Deletion Strategy}).
A visual representation of this process is depicted in Fig.~\ref{fig: point_insertion}, we randomly upsample the points to a designed number per epoch while reducing the radius of sampling on the designed epoch for coarse-to-fine optimization, which enables fast convergence during training.
Besides, we add the scale with a rendering radius which is also reduced per designed epoch for more stable convergence, and prune away invisible points that have the opacity $o_d$ in the deformed space smaller than 0.1. In Supp. Mat., we elucidate detailed values.
In the early stage of training, the coarse shape of the head can be quickly approximated through efficient deforming and rendering with an initial sparse point cloud, whereas recovering fine-grained details with dense points in the late training stage.
\subsection{Gaussian Deformation}
\label{sec: Gaussian deformation}
As the other parameters of Gaussian points are computed from their mean position, we begin by depicting each Gaussian point as standard points to clarify the deformation process.
We learn a person-specific topology-consistent template consisting of deformation blendshapes and skinning weights, similar to IMavatar~\cite{zheng2022avatar}, for expression fidelity and geometry reality such as teeth, eyeglasses, and hair.
Then, we transform the Gaussian points in canonical space to deformed space through the target FLAME expression and pose parameters based on learned blendshapes and skin weight
\begin{equation}
\label{eq:canonical_deformation_1}
    {x}_{d}={\rm LBS}({x}_{o} + {\rm B}_P(\theta;\mathcal{P}) + {\rm B}_E(\psi;\mathcal{E}), {\rm J}(\psi), \theta,  \mathcal{W})
\end{equation}
where ${\rm LBS}$ and ${\rm J}$ is the standard skinning function and joint regressor in FLAME, and ${\rm B}_P$ and ${\rm B}_E$ stand for the linear combination of blendshapes outputting pose and expression with the animation coefficients $\theta$ and $\psi$ and the blendshape bases $\mathcal{P}$ and $\mathcal{E}$.
However, maintaining these Gaussian parameters unchanged in the deformed space during the transition from canonical space to deformed space is deemed unreasonable.
Our experiments further reveal that rendering the motion of Gaussian points with static Gaussian parameters leads to undesirable blurring and inconsistent geometry in a novel pose, as demonstrated in Sec.~\ref{sec: Ablation Gaussian Deformation Fileds}.
Therefore, we model a Gaussian deformation field for animation, which outputs offsets that map the Gaussian parameters in canonical space to deformed space similar to the point deformation introduced above.
Compared with directly modeling the Gaussian parameters of each Gaussian point as an individual per-point feature, the inductive bias of MLPs enables a local smoothness prior to the deformation of parameters, similar to that of albedo~\cite{barron2011high, bergman2022generative, grosse2009ground, tappen2002recovering}.
We define our Gaussian deformation field $\mathcal{F}_{d}(\cdot)$ as
\begin{equation}
\label{eq:canonical_deformation_gaussian_parameters}
    (r_{off}, s_{off}, o_{off}, c_{off}) = \mathcal{F}_{d}(x_c, {x}_{d}, \mathcal{F}_{off}({x}_{c}))
\end{equation}
and the deformed Gaussian parameters can be updated as ${r}_{d} = {r}_{c} +  r_{off}$, and ${s}_{d} = {s}_{c} + s_{off}$, ${o}_{d} = {o}_{c} + o_{off}$, ${c}_{d} = {c}_{c} + c_{off}$.
The final deformed Gaussian $\mathcal{{G}}_{d}$ can be upadated as
\begin{equation}
\label{eq:deformed_gaussian_points}
% \begin{aligned}
    \mathcal{{G}}_{d} = \{{{x}_{d}}^{i}, {{{r}_{d}}^i}, {{{s}}_{d}}^{i}, {{{o}_{d}}^i}, {{{c}_{d}}^i}\}_{i=1:N}
% \end{aligned}
\end{equation}
\subsection{Gaussian Splatting Rendering}
\label{sec: Gaussian Splatting Rendering}
In the previous section, we obtain all the Gaussian parameters in the deformed space, allowing us to use these parameters for rendering and completing the entire training loop.
Referring to 3D-GS~\cite{kerbl20233d}, the function of Gaussian point in the deformed space is defined by the mean position ${x}_{d}$ and a covariance matrix $\Sigma$ with Gaussian equation weighted by opacity ${o}_{d}$ to explicitly represent a 3D face
\begin{equation}
\label{eq:gaussian_splatting}
    {f_{d}}^{i} (p) = {\rm sigm}({{o}_{d}}^ {i}){\rm exp}{\left(-\frac{1}{2}(p-{{x}_{d}}^{i})^{T}{\Sigma_i}^{-1}(p-{{x}_{d}}^ {i})\right)}
\end{equation}
where ${\rm sigm()}$ is the standard sigmoid function, and $\Sigma_i$ is the covariance matrix of Gaussian $i$ which can be substituted by the scale matrix $S$ and rotation matrix $R$. Additionally, these two matrices can be separated for independent optimization with a normalized quaternion $q$ for rotation and a 3D vector $s$ for scale. In our method, these correspond to ${r}_{d}$ and ${s}_{d}$ in the deformed space.
When rendering novel views, the Gaussian point needs to be projected from 3D to 2D within camera planes employing \cite{zwicker2001ewa}. 
It demonstrates that the covariance matrix ${\Sigma}^{'}$ in camera coordinates can be transformed with a viewing transformation $W$ and the Jacobian $J$ of the affine approximation of the projective transformation as follows
\begin{equation}
\label{eq:covariance_matrix}
    {\Sigma}^{'}=JWRSS^{T}R^{T}W^{T}J^{T}
\end{equation}
For differentiable rendering, we use the parameter of Gaussian point: the mean position ${x}_{d}$, the rotation ${{r}_{d}}$, the scale ${{s}_{d}}$, the opacity ${o}_{d}$, the color ${{c}_{d}}$ in the deformed space.
Subsequently, $N$ points in order of depth are overlapped a pixel to form the blending color of this pixel
\begin{equation}
\label{eq:color_blending}
    C_{\rm pix}=\sum_{i\in{\mathcal{S}}}{{{c}_{d}}^{i}}{\Pi({f_{d}}^{i}){\prod_{j=1}^{i-1}}(1-{\Pi({f_{d}}^{j}))}}
\end{equation}
where $\Pi({f_{d}}^{i})$ is the influence of each Gaussian point on the pixel, and ${\prod_{j=1}^{i-1}}(1-{\Pi({f_{d}}^{j}))}$ demonstrates the transmittance term.

\subsection{Training Objectives}
\label{sec: Network Architecture and training}
\begin{figure*}
    \centering
    \includegraphics[width=\textwidth]{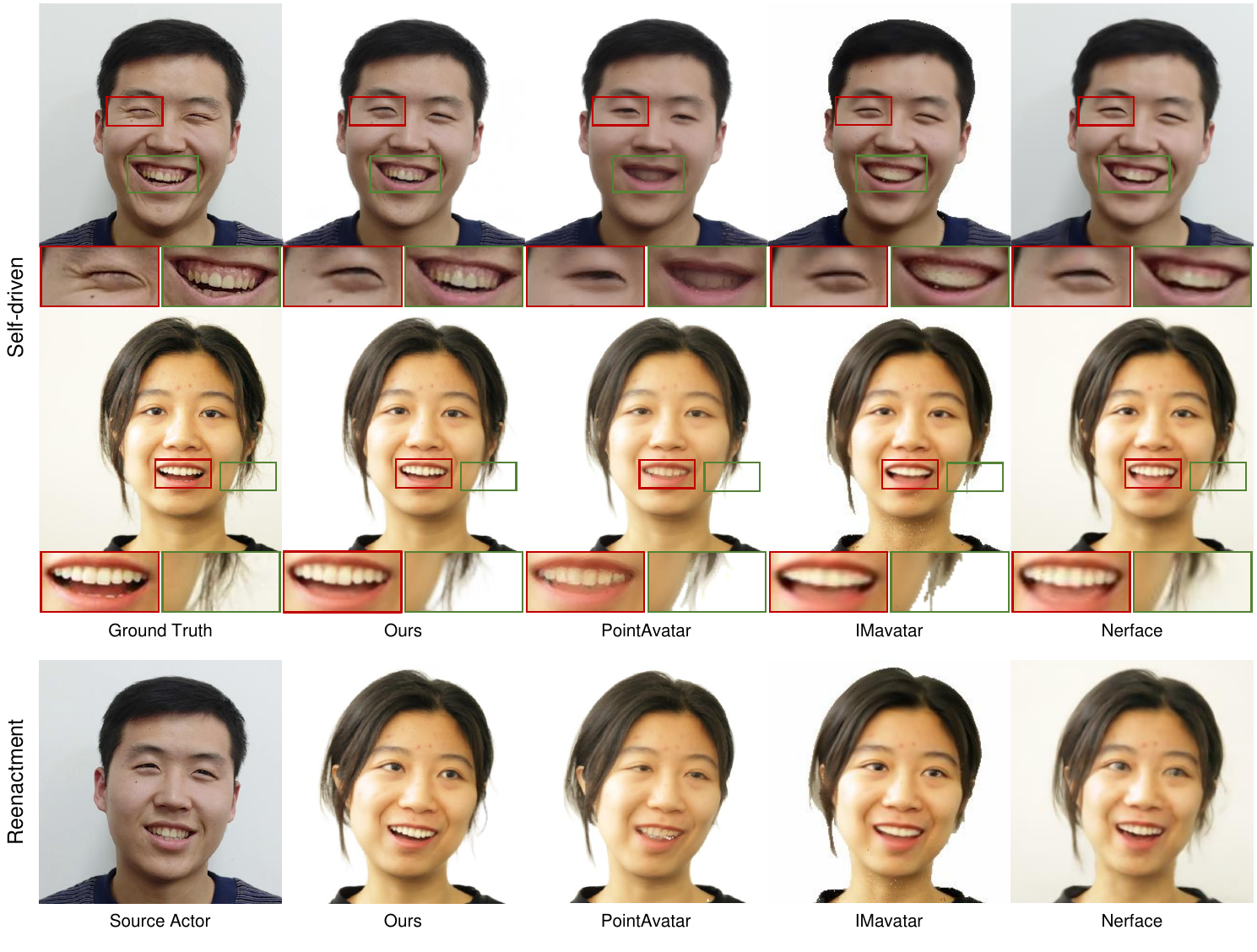}
    \vspace{-0.75cm}
    \caption{
    In qualitative comparisons, MonoGaussianAvatar demonstrates superior performance in producing photo-realistic and detailed appearances compared to state-of-the-art methods on both self-driven re-animation and facial reenactment. This is especially noticeable in the rendering of teeth details and hair textures. Furthermore, our Gaussian point-based method exhibits flexibility in capturing challenging geometries, such as thin hair strands, a capability not easily achieved by mesh-based methods.
    }
    \label{fig: comparison}
\vspace{-0.5cm}
\end{figure*}
% %
% As mentioned above, the dynamic Gaussian field is represented as two MLPs.
% %
% Specifically, we use a backbone of shared 6 fully-connected MLPs, each 256 neurons-wide followed by ReLu activation function, connected by four separate 2 fully-connected layers to predict the final colors, SDF values, and three Gaussian parameters.
% %
% The other MLP is a 4 fully-connected MLP, each 64 neurons-wide followed by ReLu activation function to predict the final deformation of the three Gaussian parameters.
%
The ultimate output of our backbone is a rendered image, and the $\textbf{RGB loss}$ serves as a supervisory signal, ensuring alignment with the per-pixel color, analogous to prior work~\cite{zheng2022avatar}
\begin{equation}
\label{eq:RGB_loss}
    \mathcal{L}_{\rm RGB}({\rm C})=\|{\rm C}-{\rm C}^{{\rm GT}}\|
\end{equation}
where ${\rm C}$ and ${\rm C}^{{\rm GT}}$ are the predicted image and ground truth image.
Besides, Gaussian splatting is an efficient point rendering that can render all the images in every training step, which makes it possible to utilize perceptual losses on the whole image. We adopt VGG~\cite{johnson2016perceptual} feature loss
\begin{equation}
\label{eq:VGG_loss}
    \mathcal{L}_{{\rm vgg}}({\rm C})=\|{\rm F}_{{\rm vgg}}({\rm C})-{\rm F}_{{\rm vgg}}({\rm C}^{{\rm GT}})\|
\end{equation}
where ${\rm F}_{{\rm vgg}}(\cdot)$ denotes the output feature of the first four layers of a pre-trained VGG network.
As well as $\textbf{RGB loss}$ and $\mathcal{L}_{{\rm vgg}}$, we adopt the loss of \cite{zheng2023pointavatar}
\begin{equation}
\label{eq:flame_loss}
\begin{aligned}
    \mathcal{L}_{{\rm flame}}&=\frac{1}{N}{\sum_{i=1}^{N}}(\lambda_e\|{\mathcal{E}_i-\widehat{\mathcal{E}}_i}\|_{2} + \lambda_p\|{\mathcal{P}_i-\widehat{\mathcal{P}}_i}\|_{2}
    \\&  +\lambda_w\|{\mathcal{W}_i-\widehat{\mathcal{W}}_i}\|_{2})
\end{aligned}
\end{equation}
where $\widehat{\mathcal{E}}$, $\widehat{\mathcal{P}}$ and $\widehat{\mathcal{W}}$ represent pseudo ground truth values, which are determined based on the nearest FLAME vertex.
Our total loss combined with a D-SSIM term is defined as follows
\begin{equation}
\label{eq:total_loss}
\begin{aligned}
    \mathcal{L} &= {\lambda_{\rm rgb}\mathcal{L}_{\rm RGB}} + {\lambda_{\rm flame}\mathcal{L}_{\rm flame}} + {\lambda_{\rm vgg}}\mathcal{L}_{\rm vgg}
    \\& + {\lambda_{\rm D-SSIM}\mathcal{L}_{\rm D-SSIM}}
\end{aligned}
\end{equation}

\newcommand{\tablequa}{
\begin{table}[t]
\small
\centering
\setlength\tabcolsep{5 pt}
% \caption{\textbf{Quantitative evaluation.} We report the quantitative results on test poses and expressions. Our method achieves better rendering quality compared to SOTA methods.}
\begin{tabular}{lcccc}

\hline  Error Metric (case 1) & L1 $\downarrow $ &  LPIPS $\downarrow$  & SSIM $\uparrow $ & PSNR $\uparrow $\\ 
\hline
Nerface{\cite{gafni2021dynamic}}  & $0.028$ & $0.1443$ & $0.8487$ & $24.05$ \\ 
IMavatar{\cite{zheng2022avatar}} & $0.028$ & $0.1622$ & $0.8454$ & $23.39$ \\ 
PointAvatar{\cite{zheng2023pointavatar}}  & $0.021$ & $0.0845$ & $0.8759$ & $25.71$ \\  
Ours  & $\textbf{0.019} $  & $\textbf{0.0733}$ & $\textbf{0.8964}$ & $\textbf{27.03}$  \\  
\hline
\hline Error Metric (case 2) & L1 $\downarrow $ &  LPIPS $\downarrow$  & SSIM $\uparrow $ & PSNR $\uparrow $\\ 
\hline
Nerface{\cite{gafni2021dynamic}}  & $0.013$ & $0.1471$ & $0.8910$ & $29.76$ \\ 
IMavatar{\cite{zheng2022avatar}} & $0.017$ & $0.1799$ & $0.9172$ & $27.43$ \\ 
PointAvatar{\cite{zheng2023pointavatar}}  & $0.015$ & $0.0798$ & $0.9211$ & $29.58$ \\  
Ours  & $\textbf{0.010} $  & $\textbf{0.0776}$ & $\textbf{0.9432}$ & $\textbf{32.55}$  \\  
\hline
\end{tabular}

\caption{\textbf{Quantitative evaluation.} We report the quantitative results on test poses and expressions. Our method achieves better rendering quality compared to SOTA methods.}
\label{tab: qua}
\vspace{-0.5cm}
\end{table}
}
\section{Experiments}
\label{sec: experiments}
\tablequa
% \begin{figure*}
%     \centering
%     \includegraphics[width=\textwidth]{figs/comparison1.pdf}
%     \vspace{-0.75cm}
%     \caption{
%     %
%     In qualitative comparisons, MonoGaussianAvatar demonstrates superior performance in producing photo-realistic and detailed appearances compared to state-of-the-art methods on both self-driven re-animation and facial reenactment. This is especially noticeable in the rendering of teeth details and hair textures. Furthermore, our Gaussian point-based method exhibits flexibility in capturing challenging geometries, such as thin hair strands, a capability not easily achieved by mesh-based methods.
%     %
%     }
%     \label{fig: comparison}
% \vspace{-0.5cm}
% \end{figure*}
\paragraph{Datasets.}
We compare our approach with state-of-the-art (SOTA) methods on one subject from IMavatar~\cite{zheng2022avatar} and one subject captured with smartphones. 
For all subjects, the same face-tracking results are applied for fair comparisons.
Besides, we utilize another subject from IMavatar~\cite{zheng2023pointavatar}, one subject from Nerface~\cite{gafni2021dynamic}, and another subject captured with smartphones to verify the contributions of our work.
\vspace{-0.7cm}
\paragraph{Baselines.}
We compare with three state-of-the-art facial reenactment methods, including 
(1) Nerface~\cite{gafni2021dynamic}, which is an implicit head avatar work leveraging dynamic neural radiance fields~\cite{mildenhall2021nerf}, 
(2) IMavatar~\cite{zheng2022avatar}, which is a 3D morphing-based implicit head avatar work with learned blendshapes from diverse expression and pose to recover detailed geometry and appearance.
(3) PointAvatar~\cite{zheng2023pointavatar}, which is built on learned blendshapes, skinning weight, and point rendering.
Together with our method, we demonstrate the photo-realism, detailed geometry, and rendering efficiency achieved by our work.

\subsection{Comparison with SOTA Methods}
\label{sec: Comparison with SOTA Methods}

\begin{figure}
    \centering
    \includegraphics[width=0.47\textwidth]{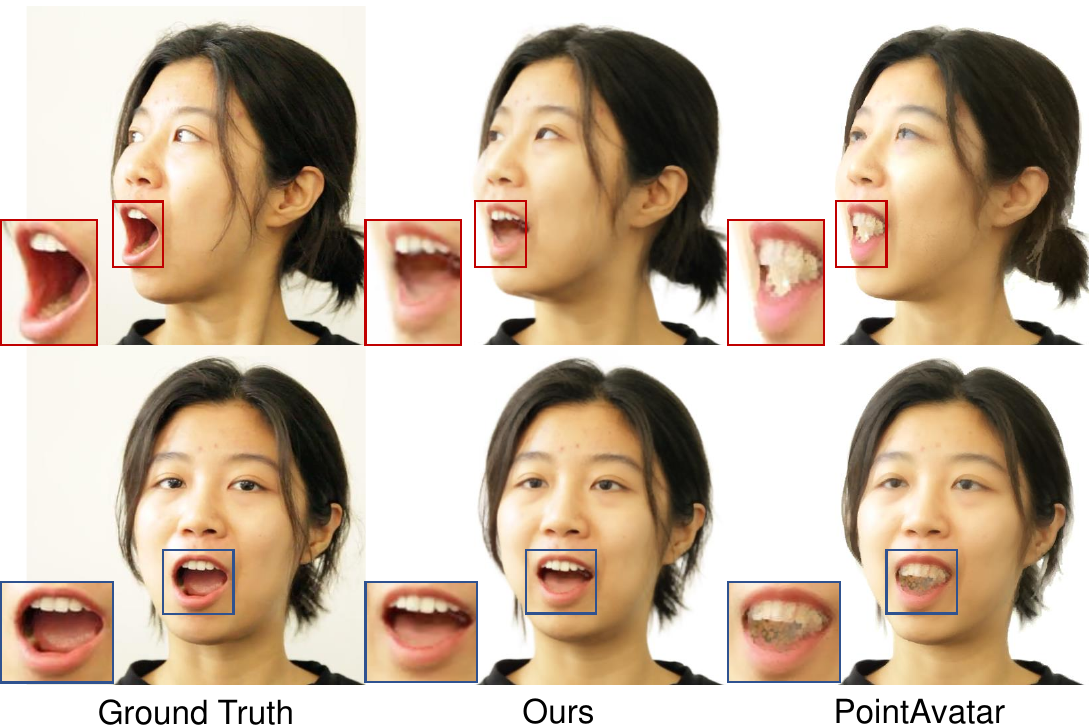}
    \caption{\textbf{Hole and Rigid structure.}
    Compare our method with PointAvatar, when the head is under substantial movement, our Gaussian points do not exhibit holes and can maintain the rigid structure of teeth.
    }
    \label{fig: hole and teeth}
% \vspace{-0.25cm}
\end{figure}
In Table~\ref{tab: qua}, we compared our method with SOTA baselines through conventional quantitative metrics (L1, LPIPS~\cite{zhang2018unreasonable}, SSIM, and PSNR).
MonoGaussianAvatar achieves the best metrics among all these SOTA methods on both DSLR videos and smartphone videos.
In the second case, our method achieves a similar metric value on LPIPS compared to PointAvatar. However, it obtains significantly higher values in the other metrics, possibly due to the limited movement in the test sequence for the second case.
\vspace{-0.5cm}
\paragraph{Teeth and detailed structures.}
In Fig.~\ref{fig: comparison}, we provide a qualitative demonstration of our method's ability to handle non-facial features introduced by rigid structures like teeth.
Compared with other methods, our method produces an avatar with finer-grained teeth.
In contrast to IMavatar, it depends on implicit fields and tends to lose details in geometry, leading to a smoother appearance of teeth.
In Fig.~\ref{fig: hole and teeth}, we further compare our method with PointAvatar in the first case, showcasing that our Gaussian points can maintain the rigidity of teeth even during substantial head movement.
\vspace{-0.4cm}
\paragraph{Hair and Skin.}
IMavatar introduces a geometry constraint on the surface, which aids in its geometry reconstruction and maintains the stability of the structure during movement. However, the implicit surface approach faces challenges in reconstructing volumetric structures, as evident in the second case where it struggles with hair details. 
Our Gaussian point-based method utilizes the point insertion and deletion strategy to dynamically allocate points in the area of hair, facilitating the effective recovery of hair geometry.
On the contrary, NerFace is a NeRF~\cite{mildenhall2021nerf}-based avatar method. Although it can model the structure of hair and recover fine-grained skin, its limited constraints in geometry may compromise rendering quality. Both cases demonstrate that our method consistently outperforms NerFace, providing more detailed skin information even in a frontal pose.
In Fig.~\ref{fig: hole and teeth}, it is also evident that our Gaussian points effectively compensate for holes when the head undergoes substantial movement.
\subsection{Ablation}

\begin{figure}
    \centering
    \includegraphics[width=0.47\textwidth]{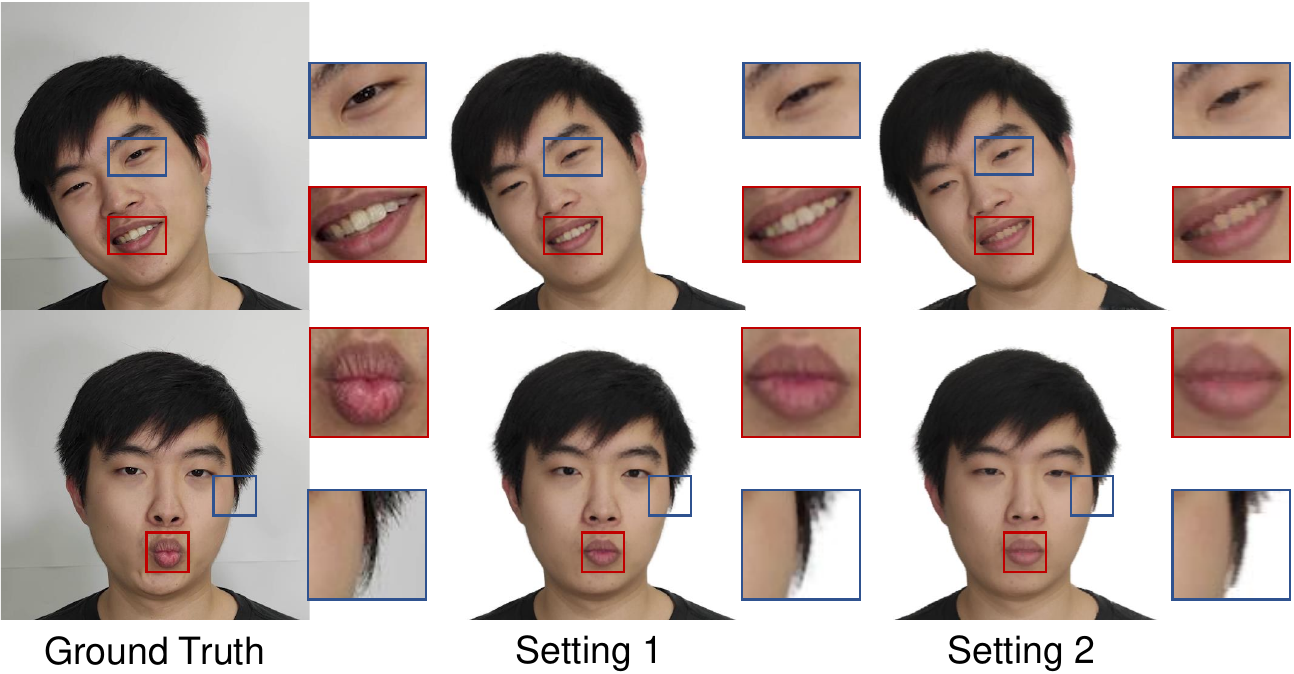}
    \caption{\textbf{Ablation: Point initial setting.}
    We explore two approaches to initialize the point setting. Initializing with a small number of points randomly sampled on a sphere enhances finer details compared to a large number of well-initialized points sampled on a FLAME template.
    }
    \label{fig: Ablation Gaussian Point Setting}
\vspace{-0.3cm}
\end{figure}

In this section, we conduct ablation experiments to investigate the impacts of Gaussian parameter deformation, the strategy of point insertion and deletion employed in our method, as well as initial setting of points.
\vspace{-0.4cm}
\paragraph{Gaussian Deformation Fileds.}
\label{sec: Ablation Gaussian Deformation Fileds}
\begin{figure}
    \centering
    \includegraphics[width=0.46\textwidth]{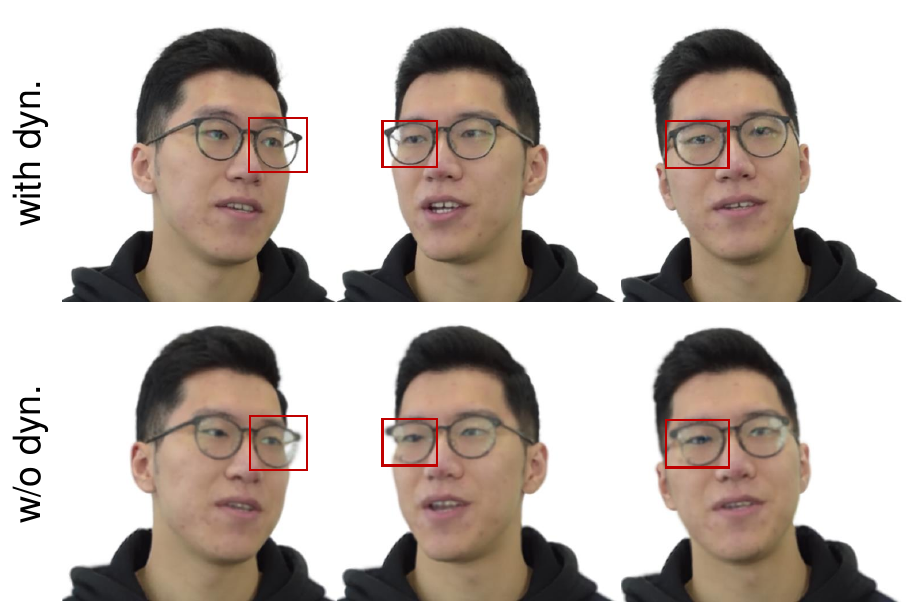}
    \caption{\textbf{Ablation: Gaussian parameter deformation.}
    The deformation field of Gaussian parameters ensures the stability of the structure of accessories, such as eyeglasses in this case, in novel poses.
    Moreover, it addresses the issue of blur when the avatar is in a novel pose.
    }
    \label{fig: Ablation Gaussian Deformation Fileds}
\vspace{-0.5cm}
\end{figure}
Our method incorporates a deformation field for Gaussian parameters in conjunction with a deformation field to describe the motions of Avatars.
%
% In the subsequent analysis, we demonstrate that the Gaussian parameters deformation field module significantly enhances the details in reenactment reconstruction.
%
Column 2 of Fig.~\ref{fig: Ablation Gaussian Deformation Fileds} reveals that as the training progresses, the incorporation of the Gaussian parameter deformation field contributes to stabilizing the structure of accessories in a novel pose. This results in stable convergence and noticeable improvements in the avatar's appearance. 
Conversely, in column 1 of Fig.~\ref{fig: Ablation Gaussian Deformation Fileds}, the absence of the Gaussian parameter deformation field leads to variations in structures in a novel pose and a blurred appearance.
\vspace{-0.4cm}
\paragraph{Point Insertion and Deletion Strategy.}
\label{sec: Ablation Point Insertion and Deletion Strategy}
\begin{figure}
    \centering
    \includegraphics[width=0.47\textwidth]{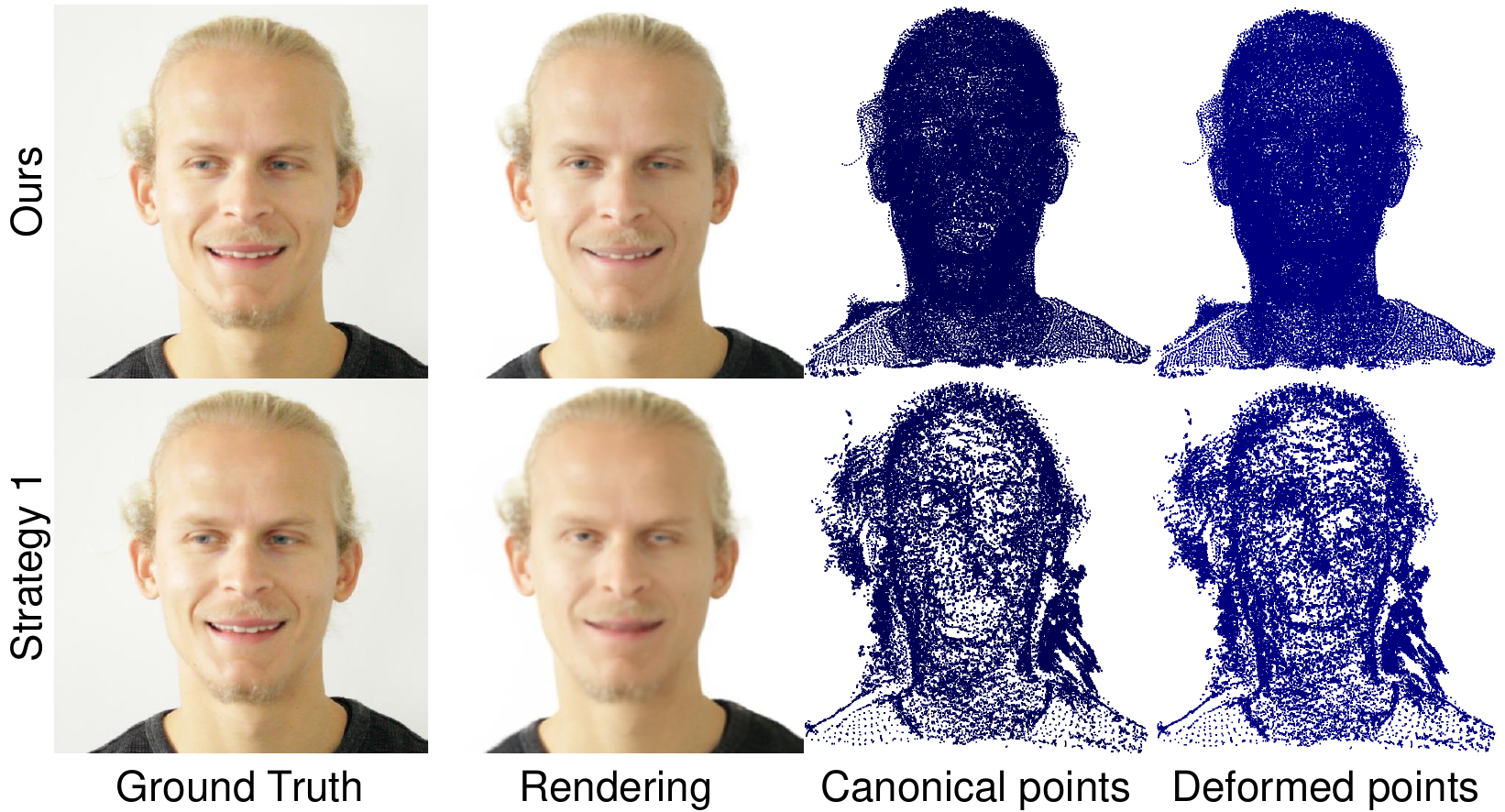}
    \caption{\textbf{Ablation: Point insertion and deletion.}
    Our strategy of point insertion and deletion enhances the geometry in both canonical and deformed spaces. Additionally, compared with the strategy of PointAvatar, the better geometry allows for more points in facial regions, resulting in the generation of finer details.
    }
    \label{fig: Ablation Point Insertion and Deletion Strategy}
\vspace{-0.4cm}
\end{figure}
%
% Our approach to point insertion and deletion shares similarities with the strategy employed in PointAvatar.
% %
% However, it's important to note that we utilize 3D Gaussian splatting as our rendering technique, and the strategy in PointAvatar can not be directly applicable to our method.
%
In Fig.~\ref{fig: Ablation Point Insertion and Deletion Strategy}, we illustrate the strategy employed in PointAvatar within our pipeline in the second column and our strategy in the first column. We display the results with the same number of points for the geometry in both canonical and deformed space, as well as the rendering appearance.
In the first column, our strategy involves pruning points with an opacity parameter smaller than 0.1 every epoch and upsampling points to reach a predetermined number within specific epochs, accompanied by a decay in sampling radius.
The second column entails pruning away points that are not the first to splat on each pixel during rendering every epoch, with a decay in rendering radius after every 5 epochs and doubling the number of points with the same sampling radius every 5 epochs.
The comparison between the two strategies clearly demonstrates the superiority in both geometry and appearance of our approach over that of PointAvatar in our specific context.
\vspace{-0.4cm}
\paragraph{Gaussian Initial Setting.}
% \label{sec: Ablation Gaussian Point Setting}
% \begin{figure}
%     \centering
%     \includegraphics[width=0.47\textwidth]{figs/setting.pdf}
%     \caption{\textbf{Ablation: Point initial setting.}
%     %
%     We explore two approaches to initialize the point setting. Initializing with a small number of points randomly sampled on a sphere enhances finer details compared to a large number of well-initialized points sampled on a FLAME template.
%     %
%     }
%     \label{fig: Ablation Gaussian Point Setting}
% \vspace{-0.3cm}
% \end{figure}
%
We tested two initial settings:
(1) As illustrated in the second row of Fig.~\ref{fig: Ablation Gaussian Point Setting}, We began by randomly sampling 400 points on a sphere to establish the initial positions of our Gaussian points. 
(2) As depicted in the third row of Fig.~\ref{fig: Ablation Gaussian Point Setting}, we sample 100000 points on the FLAME template, and optimize them with a Gaussian parameter scale in the range of 0 to 0.004.
These points are then systematically upsampled and pruned according to our point insertion and deletion strategy. 
Furthermore, we incorporate scale adjustments in the deformed space, with a fundamental rendering radius that is reduced at specific epochs.
Our experiments indicate that the first initialization method is more suitable for our work, as it consistently demonstrates superior performance and convergence in our specific context.
A potential reason is that the large number of well-initialized points can be easily identified as visual points. Therefore, point deletion becomes ineffective, leading to the refinement solely on the deformation of points.

\section{Discussion and Conclusion}
\label{sec:conclusion}

\paragraph{Limiation.} Although our approach has demonstrated its superiority compared to existing monocular 3D head avatar methods, there are still several limitations:
(1) Our current method lacks the capability to model the reflection of eyeglass lenses. This limitation could potentially be addressed by incorporating a physical model of light refraction and developing a method for disentangling the color. Additionally, the color disentangling method should not rely on normals, as Gaussian points lack the attribute of normals due to their anisotropic nature.
(2) Our method is constrained by the prior knowledge derived from the 3DMM's blendshapes and skinning weights, which limits the ability to effectively handle exceptionally challenging expressions that fall outside the scope of the pre-defined priors.
\vspace{-0.2cm}
\paragraph{Potential Social Impact.}
Since our method can reconstruct a lifelike personalized head character from a monocular video, caution should be exercised in addressing the potential misuse of the technology for creating ``deepfakes'' before its deployment.
\vspace{-0.2cm}
\paragraph{Conclusion.} We propose MonoGaussianAvatar, a Gaussian point-based explicit avatar that harnesses 3D Gaussian point representation coupled with a Gaussian deformation field applying monocular portrait videos.
Through our 3D Gaussian point representation, our method has a high-quality geometry, flexible topology, and efficient capabilities for deformation and rendering.
Moreover, by leveraging the anisotropy of our Gaussian points, our method maintains the rigidity of structures without holes, such as teeth, even under substantial movement.
%
%Moreover, it benefits from high-quality rendering results enabled by Gaussian splitting.
%
Overall, experiments demonstrate that our method outperforms other SOTA 3D head avatar approaches,
and we believe that our 3D Gaussian point representation will make progress towards effective and efficient 3D head avatar representations.

{
    \small
    \bibliographystyle{ieeenat_fullname}
    \bibliography{main}
}
% \input{sections/X_suppl}
\clearpage
\setcounter{page}{1}
\maketitlesupplementary
\newcommand{\tablerendering}{
\begin{table}[t]
\small
\centering
\setlength\tabcolsep{4 pt}
% \caption{\textbf{Quantitative evaluation.} We report the quantitative results on test poses and expressions. Our method achieves better rendering quality compared to SOTA methods.}
\begin{tabular}{lcccc}

\hline  Method & Training time (hour) & Rendering time (train)\\ 
\hline
Nerface{\cite{gafni2021dynamic}}  & $48$h & $100$s  \\ 
IMavatar{\cite{zheng2022avatar}} & $54$h & $4$s  \\ 
PointAvatar{\cite{zheng2023pointavatar}}  & $11$h & $0.1$s - $1.5$s  \\  
Ours  & $9$h  & $0.001$s   \\  
\hline
\end{tabular}

\caption{\textbf{Training and rendering time (per image).} We provide comprehensive insights into the training and rendering times of both our method and state-of-the-art (SOTA) methods. Notably, our approach attains superior efficiency in both training and rendering compared to the existing state-of-the-art methods.}
\label{tab: rendering}
% \vspace{-0.5cm}
\end{table}
}
In this supplementary document, we present additional experiments conducted with our method, specifically exploring different rendering resolutions in Sec.~\ref{sec: Different rendering resolution} and examining training and rendering efficiency in Sec.~\ref{sec: Training and Rendering Efficiency}.
Furthermore, the implementation details are provided, encompassing the strategies for point insertion and deletion in Sec.~\ref{sec: PID}, the network architecture in Sec.~\ref{sec: Network Architecture}, training details in Sec.~\ref{sec: Training Details}, evaluation details in Sec.~\ref{sec: Evaluation Details}, and the data preprocessing in Sec.~\ref{sec: Data Preprocessing}. 
For a more in-depth exploration, we recommend referring to our supplemental video.
\section{Additional Results}
\label{sec: Additional Results}
\subsection{Different rendering resolution}
\label{sec: Different rendering resolution}
\begin{figure}[H]
    \centering
    \includegraphics[width=0.47\textwidth]{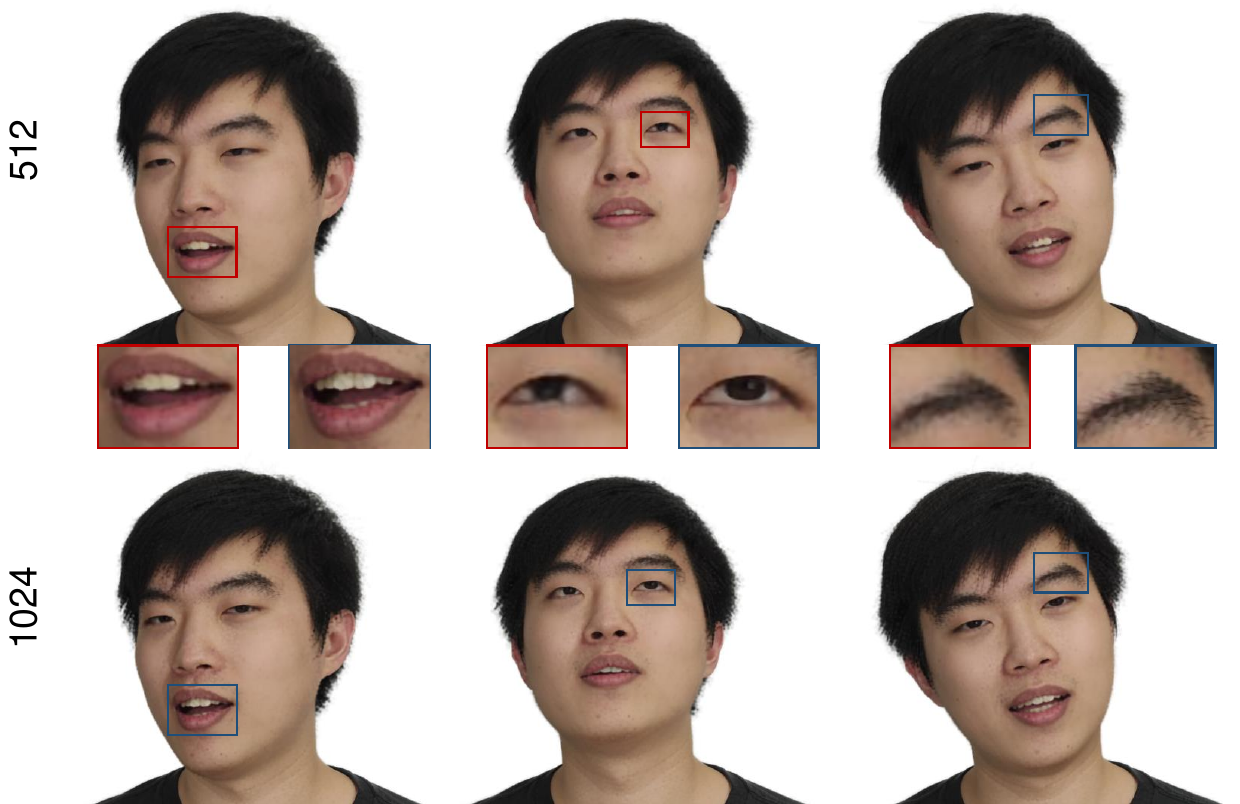}
    \caption{\textbf{Qualitative comparison of different resolutions.}
    We render our Gaussian point-based avatar representation with different resolutions. Compared to the lower resolution, the higher resolution recovers more details.
    }
    \label{fig: resolution}
% \vspace{-0.25cm}
\end{figure}

We present diverse resolution results of rendering on the same case, demonstrating the capability of our method to achieve higher resolutions with enhanced details. It is noteworthy that all the results in the main paper are rendered at a resolution of $512 \times 512$. In this section, we conduct a comparative analysis of the same case using different rendering resolutions, namely, $1024 \times 1024$ and $512 \times 512$.
As depicted in Fig.~\ref{fig: resolution}, the case rendered at a resolution of $1024 \times 1024$ exhibits superior quality in terms of appearance details.
\subsection{Training and Rendering Efficiency}
\label{sec: Training and Rendering Efficiency}
\tablerendering
\begin{figure}
    \centering
    \includegraphics[width=0.47\textwidth]{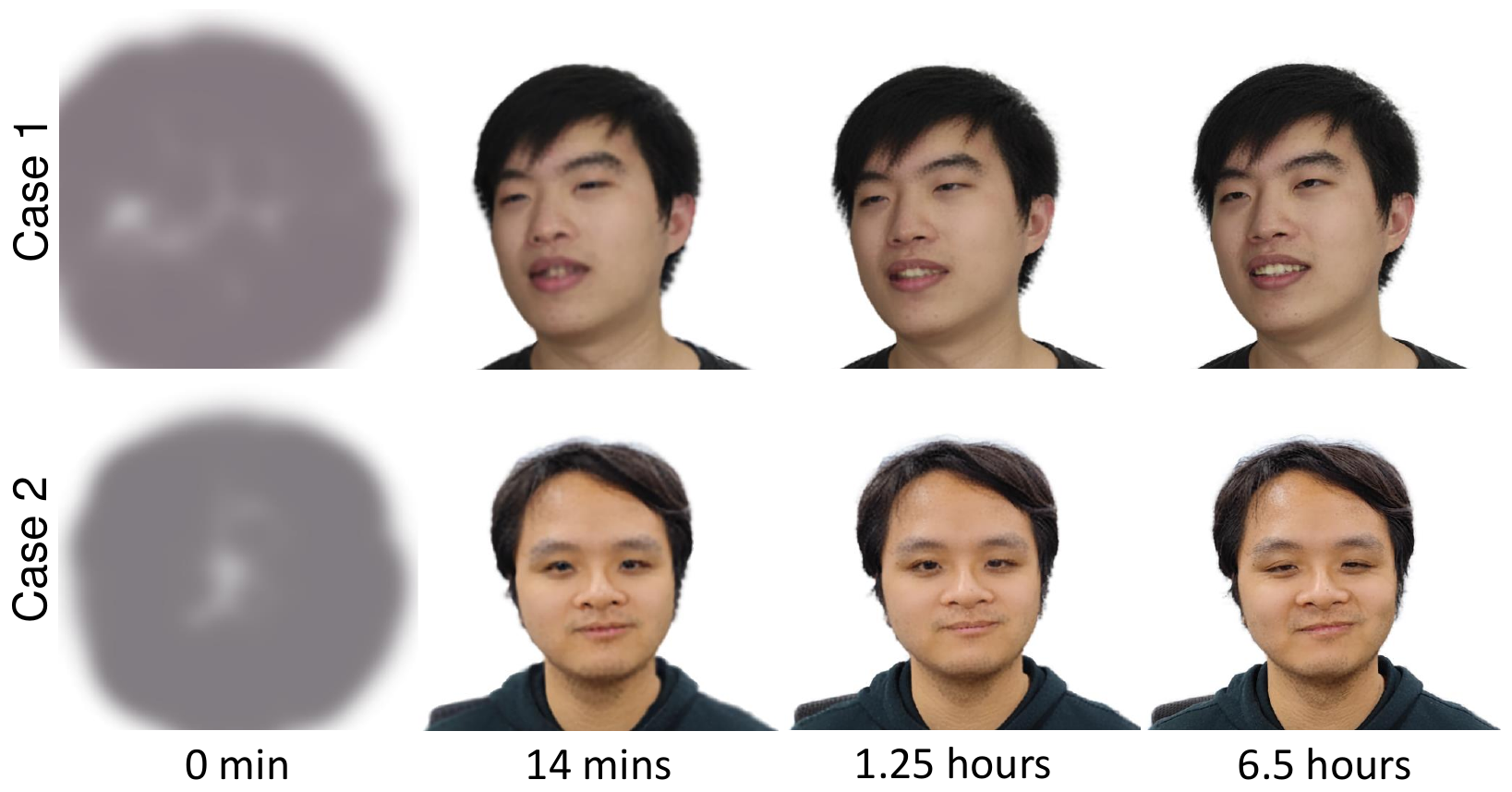}
    \caption{\textbf{Qualitative comparison of training stages.}
    We document the convergence process of our MonoGaussianAvatar. In comparison with the implicit-based method detailed in Table~\ref{tab: rendering}, our approach exhibits significantly faster convergence.
    }
    \label{fig: convergence}
% \vspace{-0.25cm}
\end{figure}
As illustrated in Table~\ref{tab: rendering}, we present a comprehensive comparison of training time and rendering time per image for the same case, underscoring the notable efficiency of our method in both training and rendering processes.
Moreover, we depict the training convergence process of our method in Fig.~\ref{fig: convergence}, illustrating its efficient training performance in two distinct cases.

\section{Implementation Details}
\label{sec: implementation details}
In this section, we provide implementation details on the strategy of point insertion and deletion, network architecture, and training details. Furthermore, our code will be made available for research purposes. It is pertinent to note that we implemented our approach in PyTorch utilizing an NVIDIA GTX 3090.
\subsection{Point Insertion and Deletion}
\label{sec: PID}
We elaborate on the detailed process of point insertion and deletion, elucidating the settings for rendering radius and sampling radius.
We randomly initialize 400 points on a sphere.
During the initial 40 epochs, we employ a two-fold strategy: pruning points with opacity below 0.1 and upsampling the points to a predetermined quantity (specified as 400, 800, 1600, 3200, 6400, 10000, 20000, 40000; the designated quantity is updated every 5 epochs). Simultaneously, the radius for both sampling and rendering is systematically reduced by a factor of $\lambda_f=0.75$ every 5 epochs.
Over the subsequent 20 epochs, we configure the designated point quantity to be 80000 and 100000, with an update occurring every 10 epochs. Additionally, the reduction in epochs for both sampling and rendering is set at 10.
During the final stage of training, we consistently upsample points to 100000 after pruning points each epoch. In the 61-100 epoch stage, both the sampling radius and rendering radius undergo a reduction by the same factor every 5 epochs. Beyond the 100th epoch, the sampling radius is maintained at a constant value of 0.004.
In our rendering process, we integrate the scales of our Gaussian points with the rendering radius.
\subsection{Network Architecture}
\label{sec: Network Architecture}
\begin{figure}
    \centering
    \includegraphics[width=0.47\textwidth]{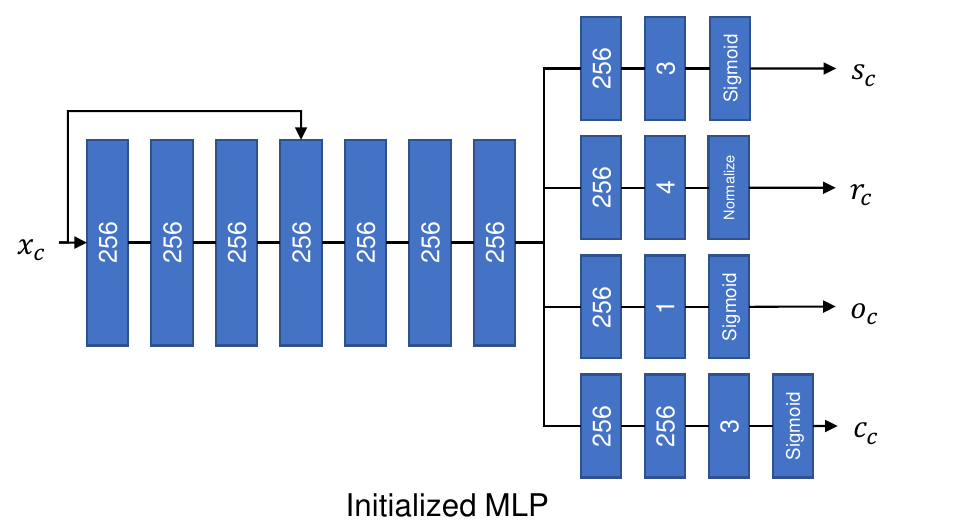}
    \caption{\textbf{The initialized MLP.}
    In the initialized MLP, each linear layer is succeeded by weight normalization, and the activation function utilized is the Softplus, with the exception of the final layer.
    }
    \label{fig: initialized}
\vspace{-0.2cm}
\end{figure}
\begin{figure}
    \centering
    \includegraphics[width=0.47\textwidth]{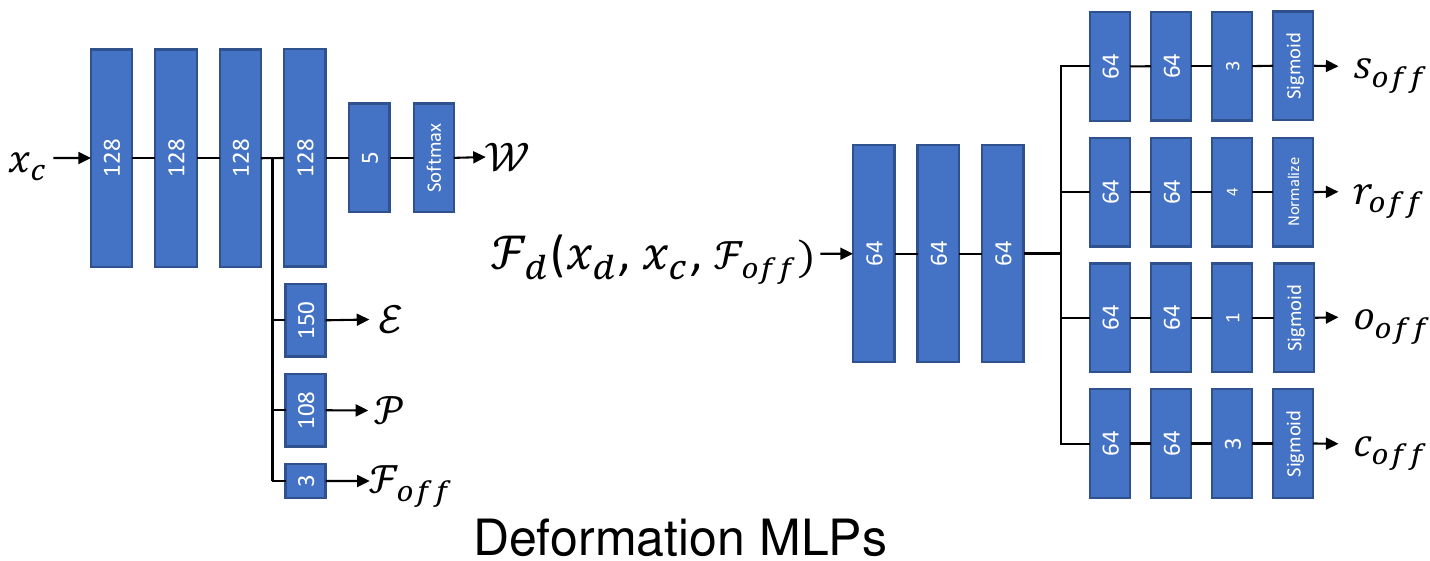}
    \caption{\textbf{The Deformation MLPs}
    In the left segment of the deformation MLP, each linear layer is succeeded by weight normalization, and the activation function utilized is the Softplus, with the exception of the final layer. Conversely, in the right segment of the deformation MLP, each linear layer is succeeded by weight normalization, and the ReLU function serves as the activation function, except for the final layer.
    }
    \label{fig: deformed}
\vspace{-0.2cm}
\end{figure}
We show the architecture of the initialized MLP in Fig.~\ref{fig: initialized} and the deformation MLPs in Fig.~\ref{fig: deformed}.
The initialized MLP, discussed in Sec.~\ref{sec: Gaussian Setting}, serves as a Gaussian parameter prediction network. Given the mean position $x_c$, it outputs the rotation $r_c$, scale $s_c$, opacity $o_c$, and color $c_c$ in the initialized space.
The left segment of the deformation MLPs, introduced in both Sec.\ref{sec: Gaussian Setting} and Sec.\ref{sec: Gaussian deformation}, delineates the motion process from the initialized space to the canonical space and ultimately to the deformed space, in terms of the mean position.
Conversely, The right segment of deformation MLPs, detailed in Sec.~\ref{sec: Gaussian deformation}, facilitates the deformation of the remaining Gaussian parameters from the initialized space to the deformed space.
\subsection{Training Details}
\label{sec: Training Details}
We show the loss weights as follows:
we choose $\lambda_{\rm RGB}=1$, $\lambda_{\rm D-SSIM}=0.25$, $\lambda_{\rm flame}=1$, and $\lambda_{\rm vgg}=0.1$ for all of our experiments. 
For the flame loss, we set $\lambda_{\rm e}=1000$, $\lambda_{\rm p}=1000$, $\lambda_{\rm w}=1$. 
The training process is optimized using the Adam optimizer with a learning rate of $lr=1e^{-4}$ and $\beta=(0.9, 0.999)$. Additionally, we implement a learning rate decay at the 80th and 100th epoch, employing a decay factor of 0.5. Moreover, we implement a decay of flame regularization at the 20th, 30th, 50th, and 70th epoch, employing a decay factor of 0.5.
\subsection{Evaluation Details}
\label{sec: Evaluation Details}
Consistent with the approach employed in NHA~\cite{grassal2022neural}, we undertake fine-tuning of pre-tracked FLMAE~\cite{li2017learning} expression and pose parameters both in the training and evaluation phases. The detailed loss weights during training are outlined in Sec.~\ref{sec: Training Details} of the Supp. Mat.
In the evaluation process, we exclusively employ the RGB loss.
\subsection{Data Preprocessing}
\label{sec: Data Preprocessing}
We adhere to the identical data preprocessing pipeline as employed in PointAvatar~\cite{zheng2023pointavatar}, which is derived from IMavatar~\cite{zheng2022avatar}. 
Additionally, we employ consistent camera and FLAME parameters across all methods. 
This ensures a fair comparison of head avatar methods, eliminating variations introduced by different face-tracking schemes during data preprocessing.
For the three human subjects captured by us, the initial preprocessing involves cropping the images to a square shape and resizing them to dimensions of both $512 \times 512$ and $1024 \times 1024$. Subsequently, we apply the data preprocessing pipeline mentioned above to further process the images
\subsection{Ethics}
\label{sec: Training Details}
We conducted experiments by capturing images of three human subjects using smartphones and additionally utilized data from three human subjects obtained from other datasets. For the 3 subjects captured by us, written consent was obtained from all subjects for the use of the captured images in this project. The data will be made publicly available for research purposes, provided that the subjects have granted permission for data publication.
%

% Having the supplementary compiled together with the main paper means that:
% % 
% \begin{itemize}
% \item The supplementary can back-reference sections of the main paper, for example, we can refer to \cref{sec:intro};
% \item The main paper can forward reference sub-sections within the supplementary explicitly (e.g. referring to a particular experiment); 
% \item When submitted to arXiv, the supplementary will already included at the end of the paper.
% \end{itemize}
% % 
% To split the supplementary pages from the main paper, you can use \href{https://support.apple.com/en-ca/guide/preview/prvw11793/mac#:~:text=Delete%20a%20page%20from%20a,or%20choose%20Edit%20%3E%20Delete).}{Preview (on macOS)}, \href{https://www.adobe.com/acrobat/how-to/delete-pages-from-pdf.html#:~:text=Choose%20%E2%80%9CTools%E2%80%9D%20%3E%20%E2%80%9COrganize,or%20pages%20from%20the%20file.}{Adobe Acrobat} (on all OSs), as well as \href{https://superuser.com/questions/517986/is-it-possible-to-delete-some-pages-of-a-pdf-document}{command line tools}.
% {
%     \small
%     \bibliographystyle{ieeenat_fullname}
%     \bibliography{main}
% }

\end{document}